\documentclass[letterpaper, 10 pt, conference]{ieeeconf}

\usepackage{amsmath}
\usepackage{amssymb}
\usepackage{subcaption}
\usepackage{xcolor}
\usepackage{graphicx}
\usepackage{lipsum}  
\usepackage{hyperref}
\usepackage{bm}
\usepackage[export]{adjustbox}

\IEEEoverridecommandlockouts
\begin{document}

\title{Independent Control of Two Magnetic Robots using External Permanent Magnets: A Feasibility Study}

\author{
        ~Joshua~Davy,~Tomas~da~Veiga,~Giovanni~Pittiglio,~\IEEEmembership{Member,~IEEE,}~James~H.~Chandler,~\IEEEmembership{Member,~IEEE}\\~and~Pietro~Valdastri,~\IEEEmembership{Fellow,~IEEE}
\thanks{Research reported in this article was supported by the Engineering and Physical Sciences Research Council (EPSRC) under grants number EP/R045291/1, EP/V009818/1 and EP/V047914/1, and by the European Research Council (ERC) under the European Union’s Horizon 2020 research and innovation programme (grant agreement No 818045). Any opinions, findings and conclusions, or recommendations expressed in this article are those of the authors and do not necessarily reflect the views of the EPSRC or the ERC.}% <-this % stops a space
\thanks{For the purpose of open access, the authors have applied a Creative Commons Attribution (CCBY) license to any Accepted Manuscript version arising. Figures \ref{fig:overview} and \ref{fig:field_lines} created with BioRender.com}
\thanks{
        ~Joshua~Davy,
        ~Tomas~da~Veiga,
        ~James~H.~Chandler,
        ~and~Pietro~Valdastri are with the STORM Lab, Institute of Autonomous Systems and Sensing (IRASS), School of Electronic and Electrical Engineering, University of Leeds, Leeds, UK.
        Email: \{\tt{el17jd, eltgdv, j.h.chandler, p.valdastri}\}@leeds.ac.uk
    }
\thanks{
        ~Giovanni~Pittiglio is with the Department of Cardiovascular
Surgery, Boston Children’s Hospital, Harvard Medical School,
Boston, MA 02115, USA. Email: \tt giovanni.pittiglio@childrens.harvard.edu
}
}
\maketitle
\thispagestyle{empty}
\pagestyle{empty}

\begin{abstract} 
 The ability to have multiple magnetic robots operate independently in the same workspace would increase the clinical potential of these systems allowing collaborative operation. In this work, we investigate the feasibility of actuating two magnetic robots operating within the same workspace using external permanent magnets. Unlike actuation systems based on pairs of electromagnetic coils, the use of multiple permanent magnets comes with the advantage of a large workspace which better suits the clinical setting. In this work, we present an optimization routine capable of generating the required poses for the external magnets in order to control the position and orientation of two magnetic robots. We show that at a distance of 15cm, minimal coupling between the magnetic robots can be achieved (3.9\% crosstalk) each embedded with 5mm diameter, 5mm length NdFeB magnets. At smaller distances, we observe that the ability to independently control the robot torques decreases, but forces can still achieve independent control even with alignment of the robots. We test our developed control system in a simulation of two magnetic robots following pre-planned trajectories in close proximity (60 mm) showing a mean positional error of 8.7 mm and mean angular error of 16.7$^\circ$.
\end{abstract}

\IEEEpeerreviewmaketitle

\captionsetup{font=small}

\section{Introduction}
\label{sec:intro}
% Surgical applications of magnetic robots MFE, MIT tentacle......
Magnetic actuation has been utilized for surgical robotic systems due to its potential for miniaturization and tetherless control \cite{kim2022magnetic}. Magnetic robots can be miniaturized to the millimeter and, more recently, the micrometer scale \cite{martin2020}, \cite{Diller2013} without loosing controllable Degrees of Freedom (DOF). Additionally, safety concerns with other robotic actuation methods (such as pressurized air) are not present with magnetic actuation \cite{van2010safety}. This ability to guide magnetic robots through natural orifices allows them to reach deep within the anatomy without the invasiveness of incisions. 

\begin{figure} 
     \centering
     \includegraphics[width=0.9\columnwidth]{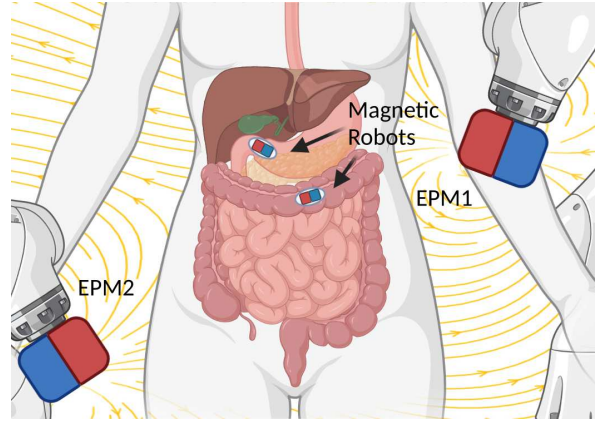}
     \caption{Overview of our proposed control scenario. We aim to independently control the full five degrees of freedom of (each internal robot each containing a single IPM). Multiple EPMs are manipulated in order to produce the desired magnetic field. The relative orientation and distance between robots affects the coupling of the robot control.}
     \label{fig:overview}
 \end{figure}

Controlling multiple magnetic robots within a shared workspace may provide new opportunities for minimally invasive surgical interventions (see Fig. \ref{fig:overview}). Magnetic robots have been equipped with surgical tools such as forceps, cameras, and biopsy needles \cite{martin2020}, \cite{forbrigger2019cable}, \cite{son2020magnetically}. The ability to have multiple magnetic robots work in collaborative operation would provide more options for surgeons in interventions by allowing multiple tools to be used in parallel. Furthermore, magnetic robots could be deployed in various parts of the anatomy (for example, along the gastrointestinal tract) and perform simultaneous microbiome sampling \cite{nam2022resonance}, \cite{shokrollahi2020blindly} or drug delivery \cite{chen2021triple}.

Independent control of magnetic robots has been extensively explored using coil-based electromagnetic systems \cite{diller2013independent}, \cite{wong2016independent}, \cite{salmanipour2018eight}, \cite{tottori2013assembly}, \cite{salehizadeh2017two}, \cite{salehizadeh2020three}, \cite{ongaro2018design}. The limited workspace of electromagnetic coil systems means the magnetic field between robots is coupled and therefore cannot be independently controlled \cite{Boehler2022}. Instead, various techniques have been employed to control the individual magnetic wrench on the robots. This has included the design of heterogeneous robots whose wrench response to the magnetic field differs, the use of underactuated control where robots are positioned and oriented with limited manipulability, and the exploitation of inter-agent magnetic interaction \cite{salmanipour2018eight}, \cite{tottori2013assembly}, \cite{salehizadeh2020three}.

By varying the external magnetic field, the resultant magnetic wrench on the robot can be controlled. This external field is usually provided by either a system of stationary magnetic coils or manipulated permanent magnets \cite{son2020magnetically}, \cite{pittiglio2022collaborative}. Actuation systems based on  electromagnetic coils have a linear relationship between applied electric currents and the generated magnetic field and can be arranged to produce fields with constant gradients across the workspace leading to a simplified control strategy. However, many designs of coil systems suffer from limited workspace size and large power supply requirements to generate strong magnetic fields \cite{abbott2020}, \cite{yousefi2021}. Alternatively, manipulated permanent magnets offer the advantage of low power requirements and large workspace but produce non-linear magnetic fields. This leads to a more complex control scenario where the relationship between magnet poses and induced wrench on the robot is non-trivial  \cite{pittiglio2022collaborative}, \cite{pittiglio2020dual}, \cite{pittiglio2022}.

Magnetic agents will experience an aligning torque with the external field and a force relating to the external magnetic field gradient. Precise control of this external field allows for a maximum of five Degrees of Freedom (DOFs) for a magnetic agent (as the torque aligned with the axis of the direction of magnetism cannot be controlled magnetically) \cite{abbott2020}. In a current-free workspace the number of DOF at a singular point in the workspace is limited to eight \cite{salehizadeh2020three},  \cite{salmanipour2020design}. Our previous work has shown how a single External Permanent Magnet (EPM) mounted to a serial manipulator is capable of independent control of all 5 DOF of a capsule robot equipped with a Internal Permanent magnet (IPM) \cite{martin2020}. Further, we have shown the independent control of eight DOFs in the case of orthogonal IPMs assuming a constant magnetic gradient across the agents with two EPMs \cite{pittiglio2020dual} \cite{pittiglio2022collaborative}. An individual EPM can fully control the five DOF of a single IPM; in this work we show that at a sufficient distance, where the interaction between each robot is small, then two EPMs could independently control the full ten DOF of two IPMs. However, as the IPMs are brought together, the inter-agent interaction between IPMs and the combined magnetic field generated by multiple EPMs affects this independence. Understanding this is key to the feasibility of operating multiple magnetic robots with full control in the same workspace, such as within the constraints of the human anatomy.

In this work, we analyse the case of attempting to control two magnetic robots each containing a single IPM operating in close proximity with EPMs (see Fig. \ref{fig:overview}). We formulate an optimization strategy capable of calculating the required poses of the EPMs for a desired wrench on the robots and further analyze the independence of control between the robots, evaluating our solutions on a dynamic simulation with a closed-loop controller based on positional and orientation feedback.

\section{Independent Actuation}
\label{sec:optimisation}
\begin{figure} 
     \centering
     \includegraphics[width=0.9\columnwidth]{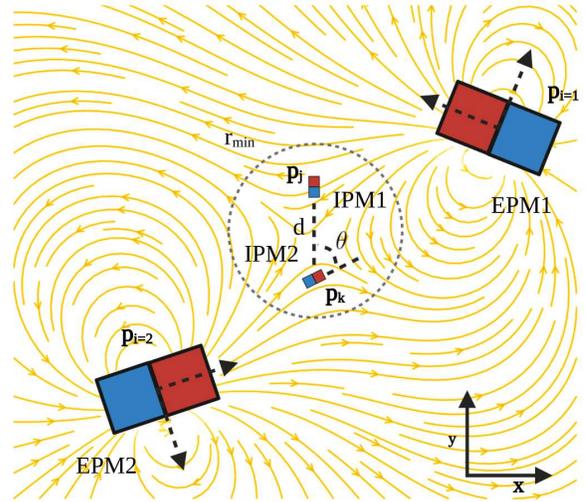}
     \caption{The combined magnetic fields of multiple EPMs create highly non-linear magnetic fields, that vary greatly in direction, magnitude and gradient across the workspace where the two IPMs are. The separation between IPMs means the local magnetic field cannot be treated linearly. Our optimization strategy instead relies on directly solving for the wrench of the IPMs.}
     \label{fig:field_lines}
 \end{figure}

\subsection{Magnetics}
We consider the control of two medical capsule robots each embedded with a single IPM actuated under the influence of $N$ EPMs. Via manipulation of the pose of the EPMs, the local magnetic field acting on the IPMs can be controlled. This local magnetic field induces a resultant magnetic wrench on the IPMs.

An IPM with magnetic moment $\mathbf{m}$  under the influence of an external magnetic field will experience a torque with an external field $\mathbf{B}$

\begin{equation}
    \boldsymbol{\tau} = \mathbf{m} \times \mathbf{B}.
    \label{torque}
\end{equation}

Further, in the case of a non-homogeneous magnetic field (i.e. a field where magnetic field gradients are present) a force is induced

\begin{equation}
    \textbf{F} = \textbf{B}_{\nabla}^T  \textbf{m},
    \label{force}
\end{equation}
where $\textbf{B}_{\nabla}$ is the spatial derivative of the magnetic field or field Jacobian

\begin{equation}
  \mathbf{B}_{   \nabla} =  \left(
\begin{array}{ccc}
    \frac{\partial B_x}{\partial x} & \frac{\partial B_x}{\partial y} & \frac{\partial B_x}{\partial z} \\
    \frac{\partial B_y}{\partial x} & \frac{\partial B_y}{\partial y} & \frac{\partial B_y}{\partial z} \\
    \frac{\partial B_z}{\partial x} & \frac{\partial B_z}{\partial y} & \frac{\partial B_z}{\partial z}
\end{array}
\right).
\end{equation}
Assuming electro-static contributions to be negligible, and therefore the workspace to be current-free, Maxwells equations state that the field Jacobian is traceless and symmetric \cite{diller2013independent}. This limits the independent components of this to five

\begin{equation}
  \mathbf{B}_{   \nabla} = 
\left(
\begin{array}{ccc}
    \frac{\partial B_x}{\partial x} & \frac{\partial B_x}{\partial y} & \frac{\partial B_x}{\partial z} \\
    \frac{\partial B_x}{\partial y} & \frac{\partial B_y}{\partial y} & \frac{\partial B_y}{\partial z} \\
    \frac{\partial B_x}{\partial z} & \frac{\partial B_y}{\partial z} & -\frac{\partial B_x}{\partial x} - \frac{\partial B_y}{\partial y}
\end{array}
\right).
\end{equation}
This, along with the three independent components of the magnetic field $\mathbf{B} \in \mathbb{R}^3$, forms the eight independent components of the magnetic field at a singular point in the workspace. In our previous work \cite{pittiglio2022collaborative}, \cite{pittiglio2020dual}, we assumed the separation of IPMs to be negligible and therefore the magnetic field and gradient to be constant across them. With greater separation of the IPMs, this assumption of linearity no longer holds.

Under the superposition principle and assuming no other source of magnetic field, the magnetic field at an IPM $j$ at position $\mathbf{p}_j$ in the workspace is the summation of the magnetic field contribution from each individual EPM and the other IPM $k$. The magnetic field at $\mathbf{p}_j$ is

\begin{equation}
    \mathbf{B}_j(\mathbf{p}_j) = \sum_{i=1}^N \mathbf{B}_i(\mathbf{m}_i,\mathbf{p}_i,\mathbf{p}_j) + \mathbf{B}_k(\mathbf{m}_k,\mathbf{p}_k,\mathbf{p}_j)
\end{equation}
where $N$ is the number of EPMs. $\mathbf{B}_i$ is the magnetic field contribution from EPM $i$ with magnetic moment $\mathbf{m}_i$ and position $\mathbf{p}_i$. (see Fig. \ref{fig:field_lines}). Equivalently $\mathbf{B}_k$ is the magnetic field contribution from the other IPM with magnetic moment $\mathbf{m}_k$ and position $\mathbf{p}_k$, with $k \neq j$. 

Equivalently the magnetic field gradient at position $\mathbf{p}_j$ is

\begin{equation}
 \mathbf{B}_{\nabla,j}(\mathbf{p}_j) = \sum_{i=1}^N  \mathbf{B}_{\nabla,i }(\mathbf{m}_i,\mathbf{p}_i,\mathbf{p}_j) + \mathbf{B}_{\nabla,k}(\mathbf{m}_k,\mathbf{p}_k,\mathbf{p}_j)
\end{equation}
where $\mathbf{B}_{\nabla,i}$ is the magnetic field gradient contribution from EPM $i$ and $\mathbf{B}_{\nabla,k}$ is the contribution from the other IPM.

From (\ref{torque}) and (\ref{force}) the induced wrench in the global reference on a IPM is 
\begin{equation}
    \mathbf{w}_j^+ = \left( \begin{matrix}
\mathbf{m}_j \times \mathbf{B}(\mathbf{p}_j)
\\ 
 \mathbf{B}_{   \nabla }(\mathbf{p}_j)^T \mathbf{m}_j
\end{matrix} \right)
\end{equation}
where $\mathbf{w_j}^+ = (\tau_{j,x}^+ \; \tau_{j,y}^+\; \tau_{j,z}^+\; F_{j,x}^+\; F_{j,y}^+\; F_{j,z}^+)^T$, represented in the global reference frame. The DOF around the axis of IPM magnetization cannot be controlled therefore, in order to analyze our system we observe the wrench on the IPMs individually, in a reference frame where the axis of magnetization aligns with the IPM $x$ axis. As in this rotated reference frame the torque around the $x$ axis will always be zero the wrench vector can be represented as the five controllable DOF for a single IPM $\mathbf{w}_j = (\tau_{j,y}\; \tau_{j,z}\; F_{j,x}\; F_{j,y}\; F_{j,z})^T$, where $j \in \{1,2\}$ is the index of the IPM.

 \subsection{Dipole Modelling}
Assuming the distances involved between EPMs and IPMs to be large enough that neglecting higher order terms is an acceptable assumption, we utilise the dipole model for modelling the magnetic fields \cite{petruska2012optimal}.
The field produced by a magnetic dipole at a point $\mathbf{p}_b$ in the workspace is 
 
 \begin{equation}
    \mathbf{B}(\mathbf{m},\mathbf{p}_a,\mathbf{p}_b) = \left ( \frac{\mu_0}{4 \pi \left \| \mathbf{r} \right \|^3}  (3\hat{\mathbf{r}}\hat{\mathbf{r}}^T - \mathbb{I} ) \right )\mathbf{m},
\end{equation}
where $\mathbf{r}$ = $\mathbf{p}_a - \mathbf{p}_b$ is the relative displacement between the dipole position $\mathbf{p}_a$ and measurement point $\mathbf{p}_b$. $\hat{\mathbf{r}}$ is the direction vector of $\mathbf{r}$, $\hat{\mathbf{r}}$ = $\frac{\mathbf{r}}{||\mathbf{r}||}$. $\mu_0$ is the vacuum permeability equal to $4 \pi \times 10^{-7}$ Hm$^{-1}$. $\mathbb{I}$ is the identity matrix.

The spatial gradient of the magnetic field produced by a dipole is 
\begin{equation}
\begin{split}
     \mathbf{B}_{\nabla}(\mathbf{m},\mathbf{p}_a,\mathbf{p}_b) =  \frac{3\mu_0}{4 \pi \left \| {\mathbf{r}} \right \|^4} ( \mathbf{m} \hat{\mathbf{r}}^T + \hat{\mathbf{r}} \mathbf{m}^T  + \\  \hat{\mathbf{r}}^T \mathbf{m}(\mathbb{I} - 5 \hat{\mathbf{r}}\hat{\mathbf{r}}^T)  )
    \end{split}
\end{equation}
\cite{abbott2020}.

\subsection{Optimization}
 Our aim is to independently control the five DOFs of both IPMs. We group these wrenches into a ten dimensional vector $\mathbf{U}$ = $(\mathbf{w}_1\; \mathbf{w}_2)^T$. For a desired $\mathbf{U}_d$ the optimum EPMs pose must be found. We formulate this as an optimization problem where we aim to minimize the norm error between desired and current wrench

\begin{equation}
\label{eq:optimization}
\underset{\mathbf{x}}{\text{argmin}}  ||\mathbf{U}_d - \mathbf{U}(\mathbf{x})||^2
\end{equation}
where $\mathbf{x}$ is the set of EPM poses which we define in terms of spherical coordinates for both position and orientation,
\begin{equation}
    \mathbf{x} = (\alpha_1\; \beta_1\; r_1\; a_1\; b_1\;...\; \alpha_N\; \beta_N\; r_N\; a_N\; b_N)^T.
\end{equation}
where $\alpha_i, \beta_i, r_i$ represent the azimuth, elevation and radius of the spherical position
\begin{equation}
    \mathbf{p}_i = \left( \begin{matrix}
x
\\
y
\\
z
 \end{matrix} \right) = r_i \left(\begin{matrix}
\cos(\beta_i)\cos(\alpha_i)
\\ 
\cos(\beta_i)\sin(\alpha_i)
\\
\sin(\beta_i)
\end{matrix}\right)
\end{equation}
and $a_i, b_i$ represents the orientation of the EPM
\begin{equation}
    \mathbf{m}_i =  ||\mathbf{m}||\left( \begin{matrix}
cos(b_i)\cos(a_i)
\\ 
\cos(b_i)\sin(a_i)
\\
\sin(b_i)
\end{matrix} \right)
\end{equation}  
where $||\mathbf{m}||$ is the magnetic moment norm of the EPM.

Compared to defining magnetic moment and position in the optimization with Cartesian coordinates, utilising polar representation reduces problem dimensionality by allowing the magnetic moment to be defined with only two variables. Further, a minimum radius around the central point of the workspace $r_{min}$, can be applied linearly rather then via additional non-linear constraints.

We utilise the \textit{interior point algorithm} with the implementation made available in the MATLAB (Mathworks, USA) programming language function \textit{fmincon}. As solutions are dependent on the initialization point, we use a random initialization from within the defined search space of EPM poses and repeat for a maximum ($M$ = 10) attempts; selecting the most optimum solution with the lowest error.

\section{System Analysis}
\label{sec:results}
\begin{figure*}[t] %% NEXT SECTION
     \centering
     \begin{subfigure}[b]{0.32\textwidth}
         \centering
         \includegraphics[width=\textwidth]{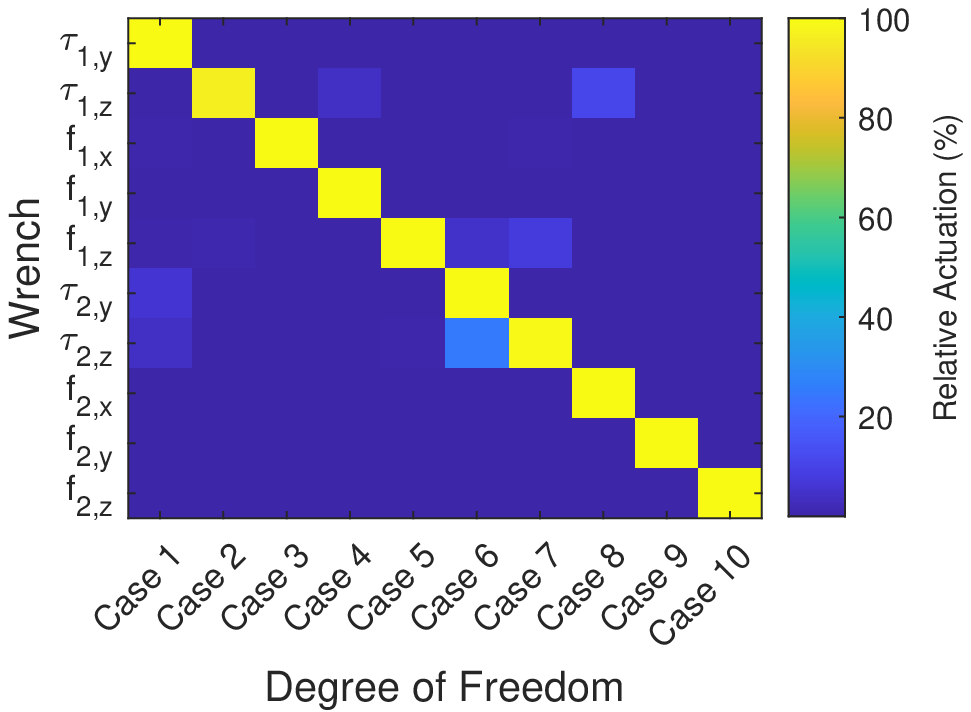}
         \caption{Activation of individual DOF with IPMs orthogonal with a separation of 15 cm ($\theta$ = $90^{\circ}$, $d$ = $0.15$ m) with two EPMS.}
         \label{fig:15_orthogonal}
     \end{subfigure}
     \hfill
     \begin{subfigure}[b]{0.32\textwidth}
         \centering
         \includegraphics[width=\textwidth]{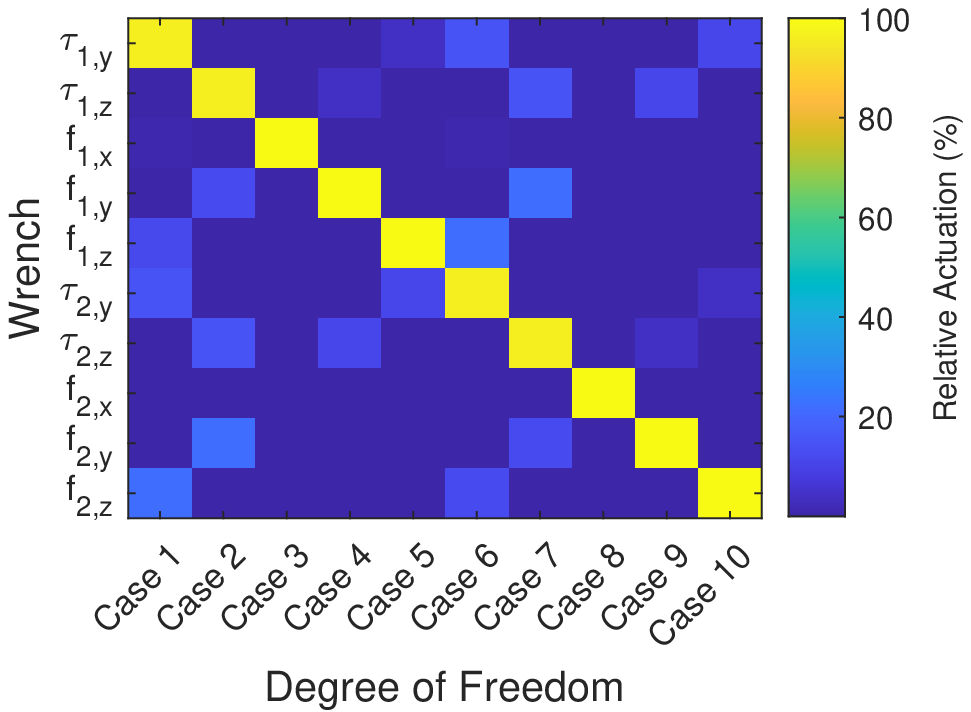}
         \caption{Activation of individual DOF with IPMs parallel with a separation of 15 cm ($\theta$ = $0^{\circ}$, $d$ = $0.15$ m) with two EPMs.}
         \label{fig:15_parallel}
     \end{subfigure}
     \hfill
     \begin{subfigure}[b]{0.32\textwidth}
         \centering
         \includegraphics[width=\textwidth]{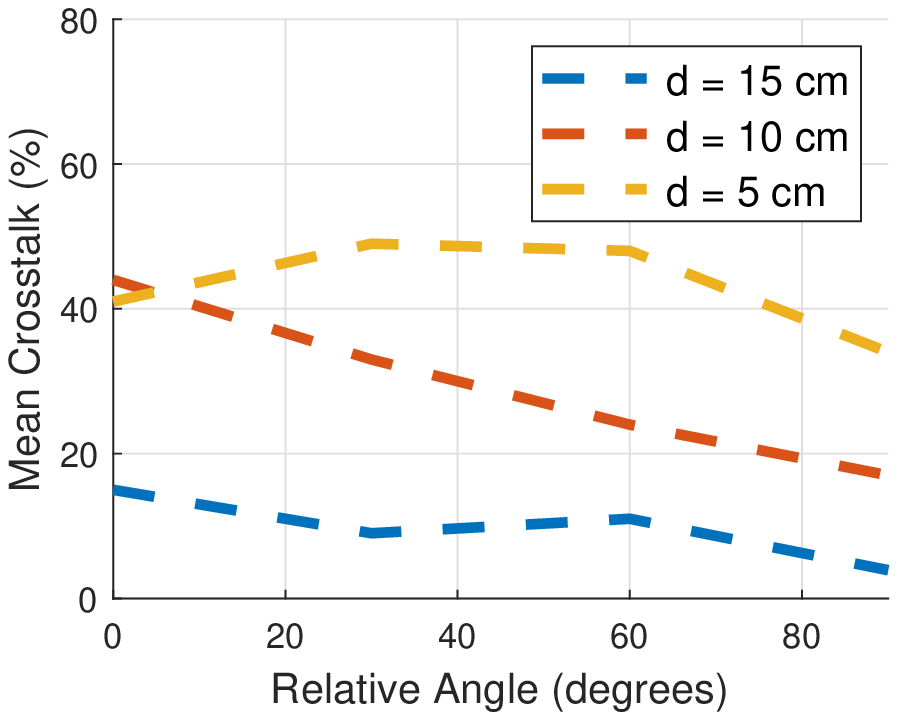}
         \caption{Mean Crosstalk across all independent activations of DOF with various seperations with increasing relative angle $\theta$.}
         \label{fig:crosstalk_distance}
     \end{subfigure}
     \hfill
        % \caption{}
        % \label{fig:15cm}
% \end{figure*}

% \begin{figure*}[t] %% NEXT SECTION
 \vskip\baselineskip
     \centering
     
     \begin{subfigure}[b]{0.32\textwidth}
         \centering
         \includegraphics[width=\textwidth]{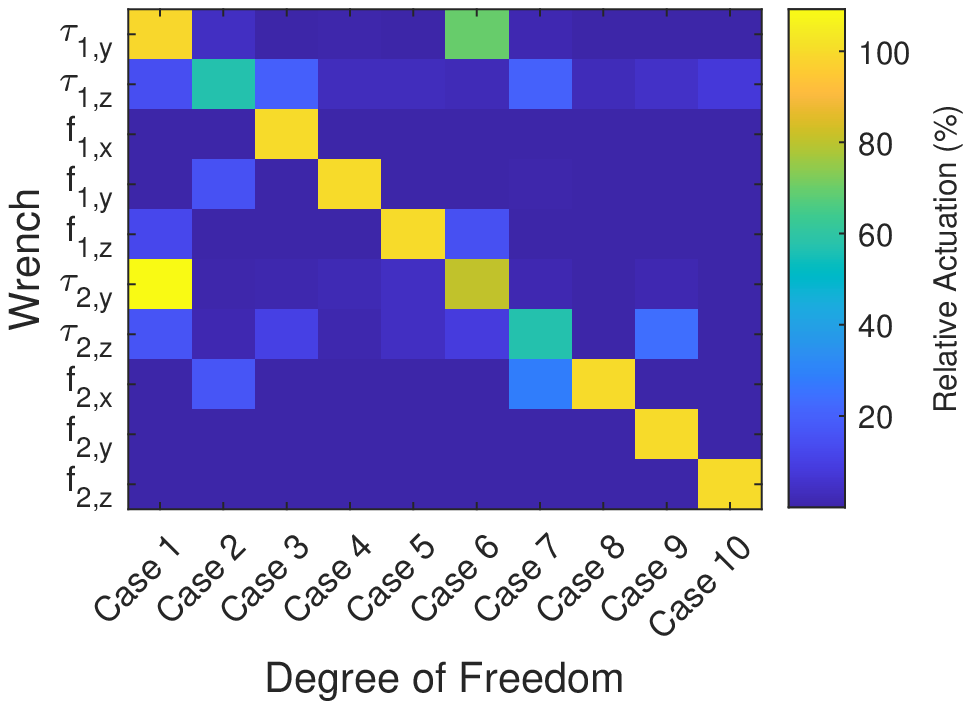}
         \caption{Activation of individual DOF with IPMs orthogonal with a separation of 5 cm ($\theta$ = $90^{\circ}$, $d$ = $0.05$ m) with two EPMs.}
         \label{fig:5_orthogonal}
     \end{subfigure}
     \hfill
     \begin{subfigure}[b]{0.32\textwidth}
         \centering
         \includegraphics[width=\textwidth]{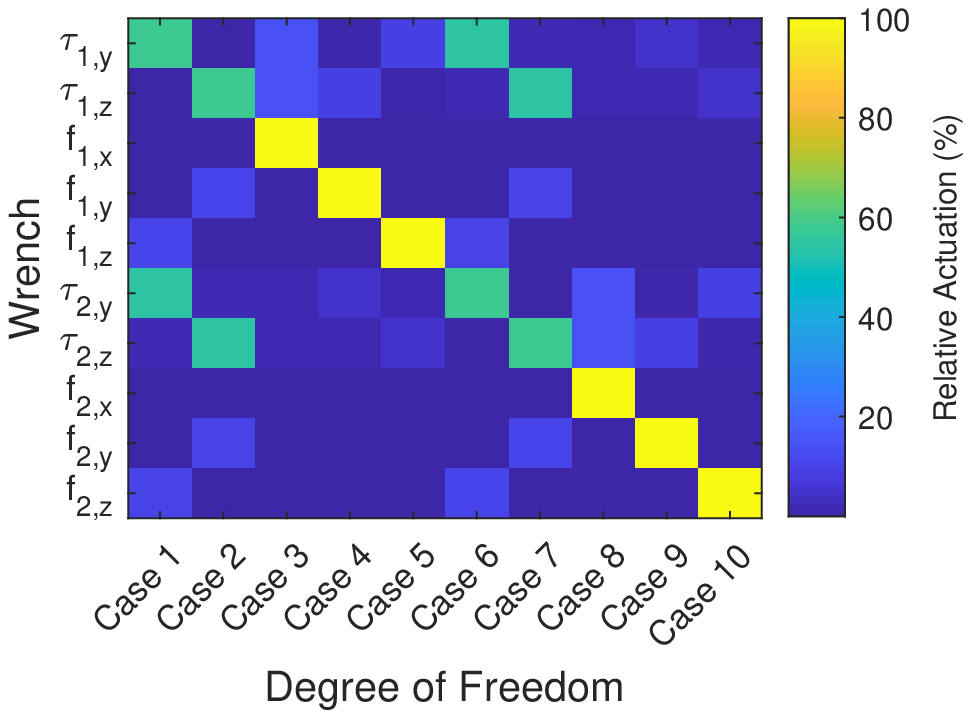}
         \caption{Activation of individual DOF with IPMs parallel with a separation of 5 cm ($\theta$ = $0^{\circ}$, $d$ = $0.05$ m) with two EPMs.}
         \label{fig:5_parallel}
     \end{subfigure}
     \hfill
     \begin{subfigure}[b]{0.32\textwidth}
         \centering
         \includegraphics[width=\textwidth]{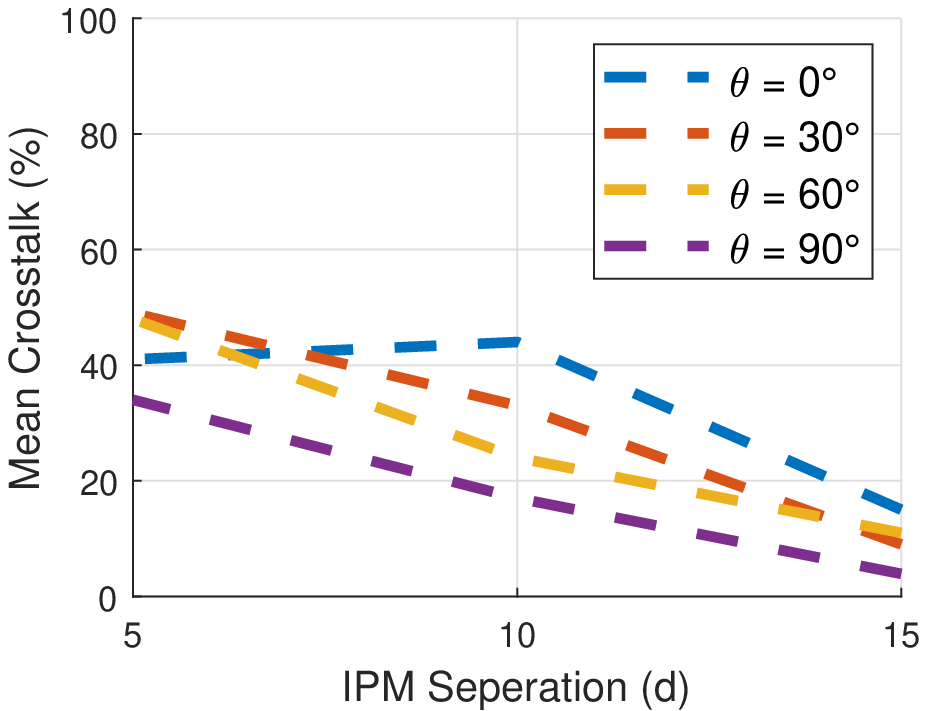}
         \caption{Mean Crosstalk across all independent activations of DOF with various relative orientations $\theta$ with increasing separation $d$.}
         \label{fig:crosstalk_angle}
     \end{subfigure}
     \hfill
        \caption{ }
        \label{fig:dualEPM}
\end{figure*}

% A discussion of suitable values
A variety of magnet sizes and shapes are utilized within magnetic robots that navigate the gastrointestinal tract. In our analysis, we utilise  cylindrical magnets made from N52 grade NdFeB  of $5$ mm diameter and $5$ mm length to represent the IPMs of the magnetic robots. By varying the distance $d$ and relative orientation $\theta$ between the IPMs we expect to see the independence of the their DOFs vary. We also consider the number of EPMs to control the system and in order to make a fair comparison we utilize a fixed volume of magnetic material split between them. We choose this to be the equivalent of the volume of the two EPMs used in our previous work ($100$ mm diameter $100$ mm length cylindrical N52 grade NdFeB magnets) \cite{pittiglio2022}. In our previous experience, these magnets represent a rough upper limit on the size of EPMs that can be manipulated with standard collaborative manipulators without causing electromagnetic interface with the manipulator joint encoders. The distance between EPMs is constrained in order to prevent their mutual attraction overpowering the manipulators. To do so, we set the minimal distance between EPMs to ensure their mutual attraction does not exceed $10g$ N, which represents a typical maximum payload of standard collaborative manipulators $\approx 10$ kg ($g$ = $9.81$ ms$^{-2}$). We also impose a minimum radius $r_{min}$ of $0.15$ m between EPMs and the center of the IPMs (see Fig. \ref{fig:field_lines}).
We aim to independently actuate the forces and torques on each IPM to values of $0.5$ N and $0.05$ Nm$^{-1}$ respectively whilst minimizing the activation of other DOFs. The required force and torques are application dependent but these represent typical values that could manipulate a magnetic robot while overcoming external forces such as gravity or tether drag \cite{scaglioni2019explicit}, \cite{pittiglio2019magnetic}.

\begin{figure*}[t] %% NEXT SECTION
     \centering
     \begin{subfigure}[b]{0.32\textwidth}
         \centering
         \includegraphics[width=\textwidth]{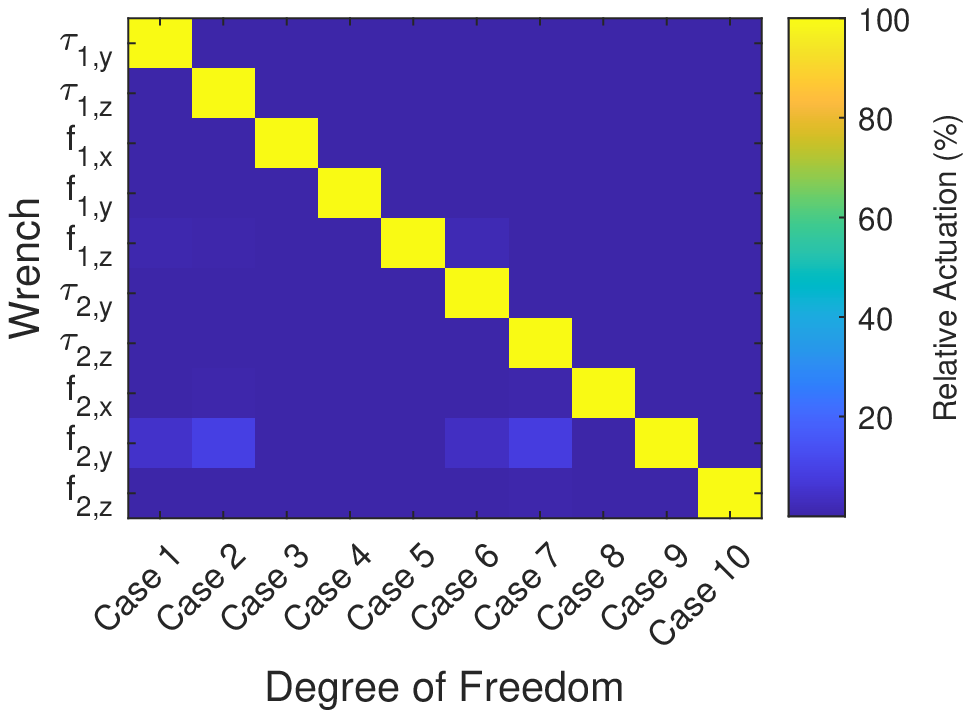}
         \caption{Activation of individual DOF with IPMs parallel with a separation of 15 cm ($\theta$ = $90^{\circ}$, $d$ = $0.15$ m) with three EPMS.}
         \label{fig:tri_15}
     \end{subfigure}
     \hfill
     \begin{subfigure}[b]{0.32\textwidth}
         \centering
         \includegraphics[width=\textwidth]{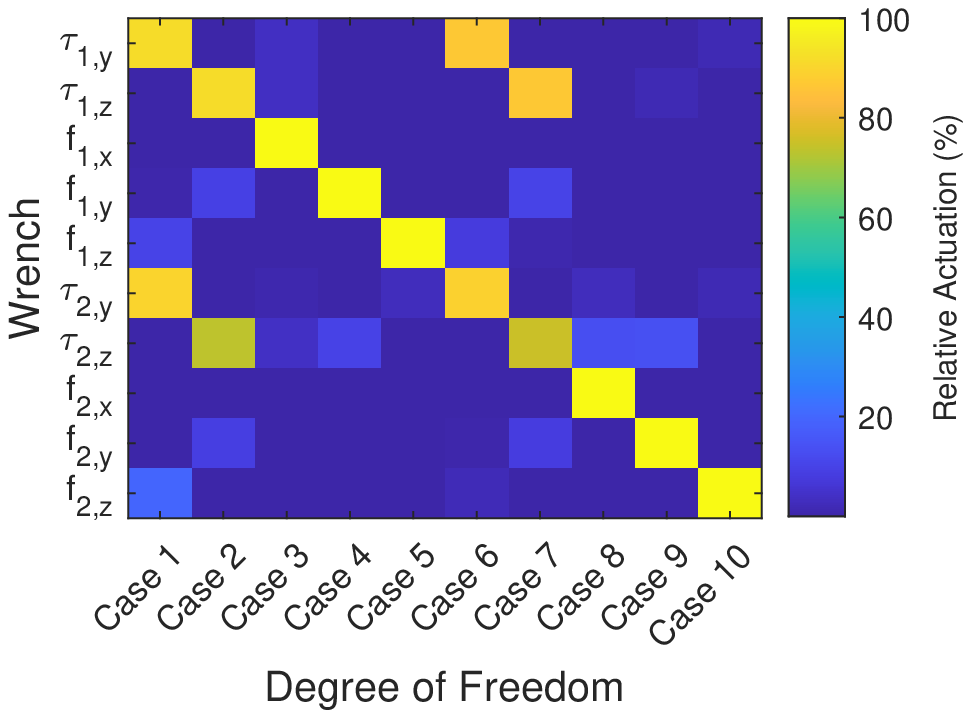}
         \caption{Activation of individual DOF with IPMs parallel with a separation of 5 cm ($\theta$ = $0^{\circ}$, $d$ = $0.15$ m) with three EPMs.}
         \label{fig:tri_5}
     \end{subfigure}
     \hfill
     \begin{subfigure}[b]{0.3\textwidth}
         \centering
         \includegraphics[width=\textwidth]{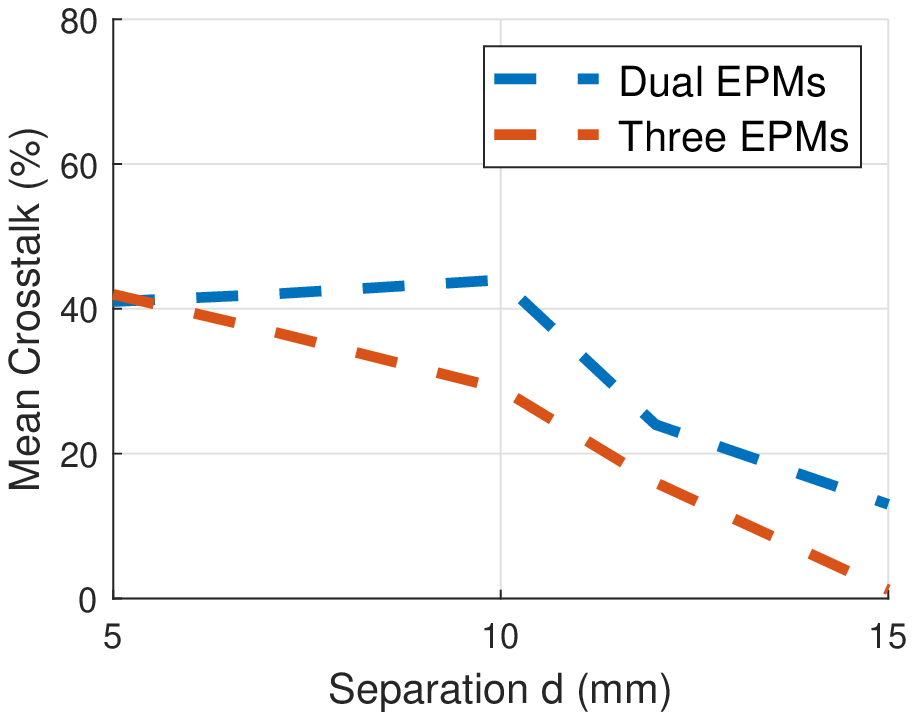}
         \caption{Mean crosstalk across all independent activations of DOF for parallel IPMs compared between dual and three EPMs.}
         \label{fig:compare}
     \end{subfigure}
     \hfill
        \caption{ }
        \label{fig:triEPM}
\end{figure*}

\subsection{Two EPMs}
We first consider the manipulation of two IPMs using two EPMs. Here, the number of control variables (i.e. the DOF of EPM poses, 10) matches the number of output variables. As stated above, with significant separation of the IPMs, their control can be considered separately with an individual EPM manipulating a single IPM. This scenario would lead to full control. Once this separation is decreased, then this independence decreases. For independence of control, the best case scenario for relative orientation of IPMs is if they are orthogonal ($\theta$ = 90$^{\circ}$). From our previous works, we have shown this provides a minimum of eight DOF with zero separation \cite{pittiglio2020dual}. If this condition was applied for the case of parallel IPMs (where the resultant magnetic field and gradient vector is the same across them), this would only lead to 5 independent DOFs with the actuation between IPMs fully coupled. Instead, we make no assumption of constant gradient leading to this greater DOF. 

To observe this independence, we utilize a metric to define the ability to independently actuate a single DOF. We define crosstalk as

\begin{equation}
     C(\mathbf{U}, i) = \max_{j  \in \{1, 10\}} \frac{|U_j|}{|U_i|}, \text{ }i \neq j.
\end{equation}
where $i$ is the index of the desired independently actuated component of $\mathbf{U}$. A crosstalk of 0\% would show that the configuration of the EPMs can individually activate $U_i$ without any activation of any other component.

To evaluate the ability to independently actuate all ten DOF of the two IPMs using two EPMs, we utilize our presented optimization strategy to find, if given the constraints on EPM poses, each DOF that can be activated with minimal activation of the others. Fig. \ref{fig:15_orthogonal} and \ref{fig:15_parallel} show the relative activation of each DOF for the case of orthogonal and parallel IPMs with a separation of 15 cm ($d$ = $0.15$ m) and result in small mean crosstalks of 3.9\% and 15.0\% respectively. Fig. \ref{fig:5_orthogonal} and \ref{fig:5_parallel} show the same case of orthogonal and parallel IPMs but with a reduced separation of 5 cm ($d$ = $0.05$ m). As expected, the independence of control is lost with a smaller distance between IPMs. In particular, it can be observed in the parallel case, tight coupling with the torques of the IPMs with crosstalk reaching 100\%. This suggests that two EPMs cannot create largely different directions of magnetic fields over short distances.
The forces however, (which are proportional to the field gradients), have much less coupling. Fig. \ref{fig:crosstalk_distance} shows how with greater IPM separation, the mean crosstalk decreases. 
 Fig. \ref{fig:crosstalk_angle} shows how the relative IPM angle affects the maximum crosstalk when independently activating each DOF. It can be observed, that the greater angle leads to greater independence of activation.

\subsection{Three EPMs}
The use of more but smaller EPMs comes with several advantages for independent control. The first is the ability to create more complex magnetic fields because of the greater number of sources of magnetic fields. Secondly, the mutual attraction between EPMs is decreased allowing them to be operating in closer proximity safely. However, with more EPMs the number of control variables increases. This complicates the control, but may come with the advantage of redundancy due to multiple solutions. Two EPMs are enough to completely control the 8 magnetic DOF of a singular point in the workspace, but with greater separation, the use of three EPMs allows us to control more DOF across multiple points in the workspace.
Fig. \ref{fig:tri_15} shows the case of parallel IPMs with a separation of 15 cm ($\theta$ = 0$^{\circ}$ $d$ = $0.15$ m) for the three EPM case achieving a crosstalk of just 1.6\% compared to 15.0\% of the dual EPM case. However, at small separations, the same issue of torque coupling is apparent (see Fig. \ref{fig:tri_5}). Fig. \ref{fig:compare} compares the separation with crosstalk for both the dual and three EPM cases.

\section{Dynamic Simulation}
\label{sec:simulation}

\begin{figure}[b]
    \centering
    \includegraphics[width=\columnwidth]{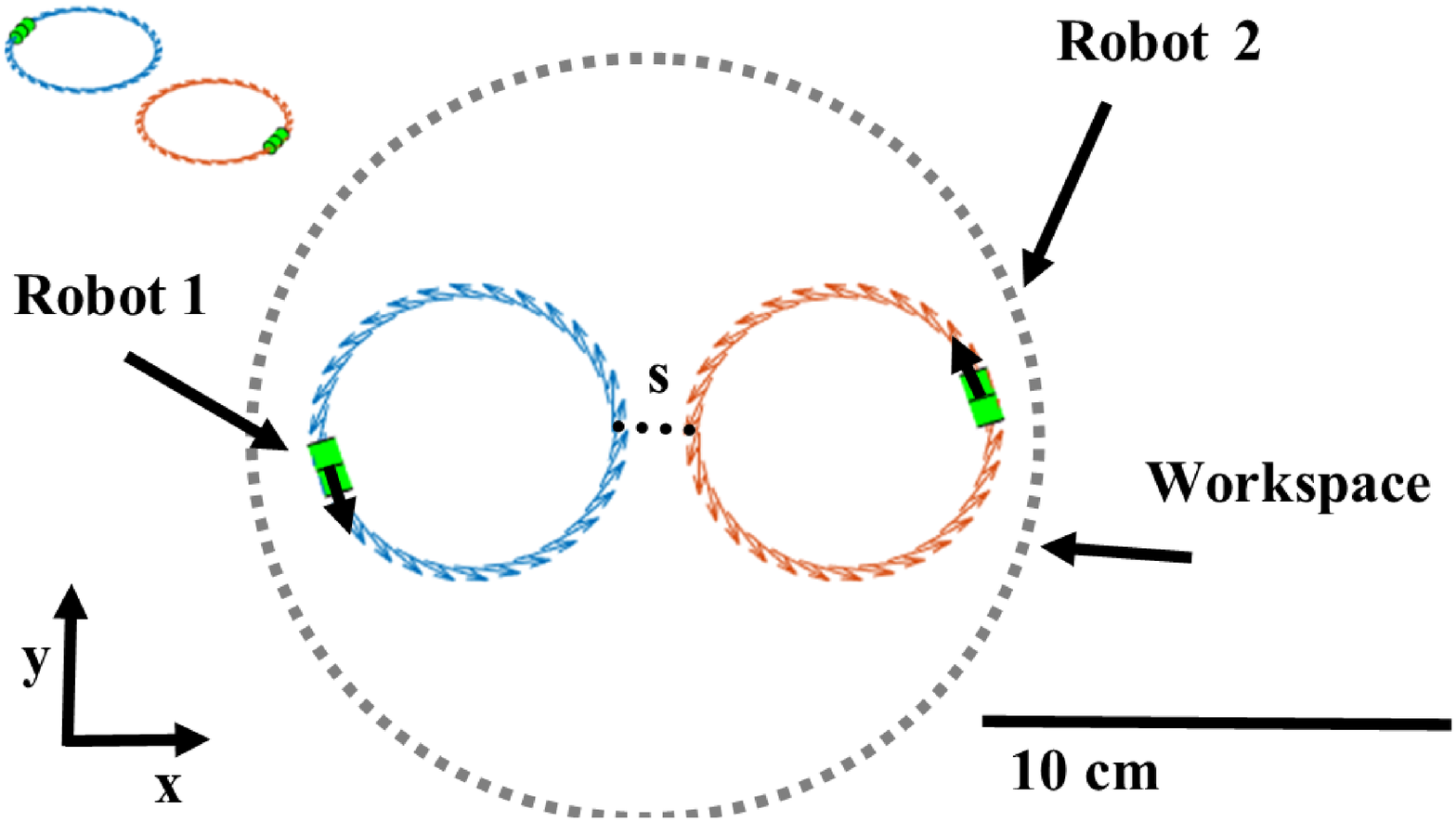}
    \caption{Overview of our simulation setup. Two magnetic robots each embedded with a singular IPM are controlled through pre-planned trajectories within the workspace. The controller positions the EPMs in order to give the desired wrench on the robots.}
    \label{fig:simulation}
\end{figure}

\begin{figure*}[t] %% NEXT SECTION
     \centering
     \begin{subfigure}[t]{0.32\textwidth}
         \centering
         \includegraphics[width=\textwidth]{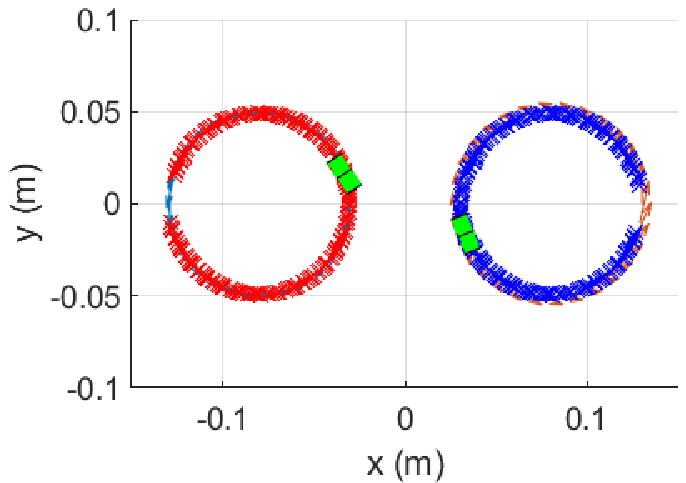}
         \caption{Robot trajectories with a minimal separation of 60mm. The controller maintains stability throughout the entire trajectory. Red crosses correspond to robot 1's path. Blue crosses correspond to robot 2's path.}
         \label{fig:sim_succsess}
     \end{subfigure}
     \hfill
     \begin{subfigure}[t]{0.32\textwidth}
         \centering
         \includegraphics[width=\textwidth]{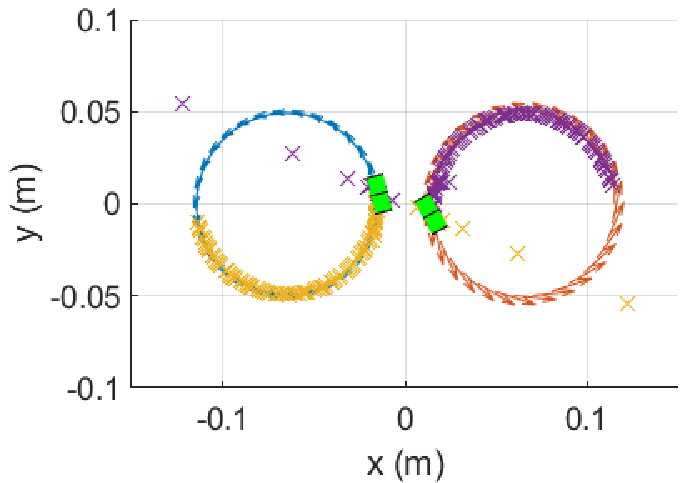}
         \caption{Robot trajectories with a minimal separation of 30mm. As the robots enter close proximity, the controller becomes instable and the robots diverge from the path.  Yellow crosses correspond to robot 1's path. Purple crosses correspond to robot 2's path.}
         \label{fig:sim_fail}
     \end{subfigure}
     \hfill
        \begin{subfigure}[t]{0.32\textwidth}
         \centering
         \includegraphics[width=\textwidth]{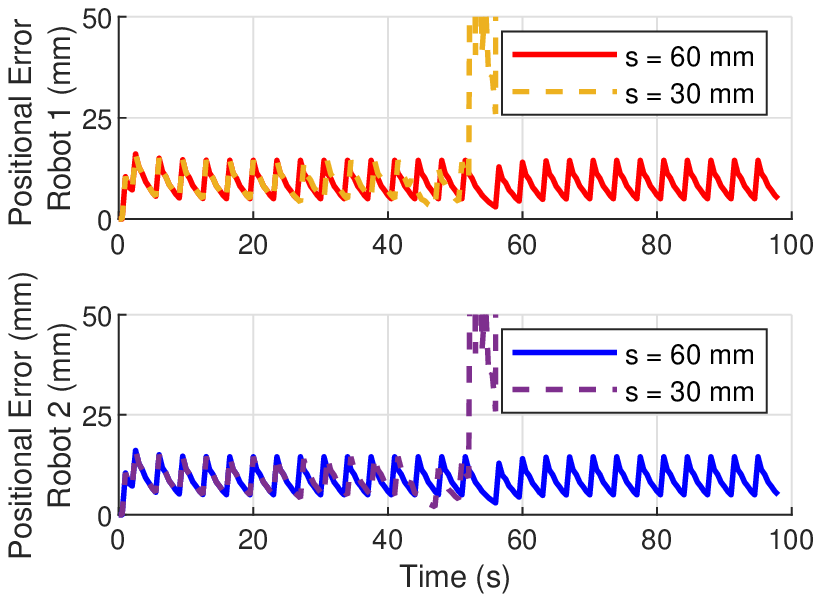}
         \caption{Positional error for both robots in both scenarios. In the second scenario, at $t$ = $51$ s, control is lost when the robots enter close proximity ($33$ mm).}
         \label{fig:sim_error}
     \end{subfigure} 
     
        \caption{ }
        \label{fig:sim}
\end{figure*}

%Overview
In order to validate the ability to control two magnetic robots in close proximity with multiple EPMs, we utilize a dynamic planar simulation of two robots each embedded with a singular IPM. By manipulating the pose of the EPMs in the simulation, the magnetic wrench on IPMs can be controlled. The IPMs embedded in the robots are the same cylindrical NdFeB magnets used in our above analysis ($5$ mm diameter, $5$ mm length). Although it was observed that three EPMs lead to better independence of control, we choose to focus this simulation analysis on the case of two EPMs due to the simpler control scenario and being more applicable to our previous work based on real-world dual EPM actuation \cite{pittiglio2022collaborative}. As before, the EPMs are cylindrical NdFeB magnets ($100$ mm diameter, $100$ mm length) and the robots operate in a spherical workspace with radius $r_{min}$ = $15$ cm.

% COntrolller
Pre-planned trajectories for the robots are formulated within the central workspace. These trajectories are formed of sets of target poses for the robot at a given time-step. A PID controller calculates the required wrench on each robot in order to reduce the error between the current pose and target pose at a frequency of 2Hz. In our simulation, we presume the EPMs can be moved instantaneously between poses. This simplification would not be applicable to real world application, where the transition of EPMs will affect the wrench on the robots, but for the purpose of this feasibility study suffices for understanding if control of multiple magnetic robots is possible with EPMs. Fig. \ref{fig:simulation} visualizes the robots and the pre-planned trajectories. The inter-agent wrench between IPMs is modelled in the simulation. The pre-planned trajectories are chosen in order to have the robots start with a significant distance between them (in order to ensure their control can be separated) but come to a minimal separation $s$ (see Fig. \ref{fig:simulation}). Videos showing the robot trajectories and the corresponding EPM poses can be found in the Supplementary Video.

To compare the effect of separation on the control of the robots, we alter the trajectories to control the separation $s$. In the first scenario (Fig. \ref{fig:sim_succsess}), the minimal separation is $60$ mm. The controller is capable of maintaining control of the robot poses throughout the trajectory with a mean positional error of 8.7mm and mean angular error of $16.7^\circ$. Although from our above analysis, at these small separations, there is large crosstalk between applied wrench, the smooth trajectories used here require small wrenches on the robot to follow and therefore control is maintained.

The second scenario (Fig. \ref{fig:sim_fail}) shows a reduced minimal separation of $30$ mm. At $t$ = $51$ s, the robots enter into close proximity ($33$ mm), and the controller fails (Fig. \ref{fig:sim_error}). This is due, to the inability of the optimizer to find suitable EPM poses that can generate the required wrench on the robots. At this short distance the inter-agent wrench is large, and the inability to create large change in magnetic field properties over the short distance means that the wrench between robots cannot be sufficiently controlled independently. In reality, this would lead to a instability where the robots would attract each other and collide. 

\section{Conclusion}
\label{sec:conclusion}
In this work, we have considered the ability to control two magnetic robots within a shared workspace with multiple manipulated EPMs. We have developed an optimization scheme capable of generating the optimum pose for EPMs in order to induce a desired magnetic wrench onto the robots. It can be observed that with separation between robots ($15$ cm), full independence of control ($3.9$\% crosstalk) can be achieved; however, with smaller distances and alignment, the coupling between the torques on both robots increases. We have analyzed how, the relative orientation and separation affects this coupling and how it could be potentially countered by using more EPMs. Finally, we have utilized our optimization strategy as part of a dynamic controller in order to control two simulated magnetic robots in varying proximity on pre-planned trajectories. Our simulation confirms our analysis, showing that at small separations the coupling between robots leads to instability in our control. Future work will consider how these robots can be utilised to operate together to perform collaborative operation or operate separately in various parts of the gastrointestinal tract.

In our control scenario, we have assumed that EPMs can be moved instantly between time steps. In reality, the transition between poses will affect the resultant wrench on IPMs and must be considered.  The stability of our scenario has not been considered and a better understanding of how accumulating errors will affect the control must be considered for real world feasibility.
Future work, would consider how collaborative manipulators could be utilized for controlling the EPMs. This would impose tighter and more complex constraints on potential EPM poses due to the reach of the manipulators. Currently, there is no constraint between sequential EPM poses, this should be constrained in order to produce realistic EPM trajectories, this may limit controllability.
The tight coupling between magnetic torques at small distances limits the controllability with small separations. Differences in geometric or magnetic design between magnetic robots could be utilized to better separate their response to similar magnetic fields. The ability to control the robots is dependent on receiving good feedback of their relative position to the EPMs; future work should consider how this feedback can be obtained using magnetic field or imaging sensors.

% \ifCLASSOPTIONcaptionsoff
%   \newpage
% \fi

\bibliographystyle{IEEEtran} 
\bibliography{references}{}

\end{document}